\documentclass[10pt,conference]{IEEEtran}
\IEEEoverridecommandlockouts
\usepackage{cite}
\usepackage{amsmath,amssymb,amsfonts,bm}
\usepackage{algorithmic}
\usepackage{flushend}
\usepackage[table,xcdraw]{xcolor}
\usepackage{graphicx}
\usepackage{multirow}
\usepackage{comment}
\usepackage{url}

\usepackage[linesnumbered,ruled,noend]{algorithm2e}

\SetAlgoCaptionSeparator{:}
\SetNlSty{texttt}{}{:}
\usepackage[T1]{fontenc}
\usepackage[font=small,labelfont=bf,tableposition=top]{caption}

\DeclareCaptionLabelFormat{andtable}{#1~#2  \&  \tablename~\thetable}
\SetAlgoCaptionSeparator{:}
\SetNlSty{texttt}{}{:}

\newcommand{\flowminer}{\emph{FlowMiner}\xspace}

\newcommand{\texada}{\emph{Texada}\xspace}
\newtheorem{definition}{Definition}[section]

\def\BibTeX{{\rm B\kern-.05em{\sc i\kern-.025em b}\kern-.08em
    T\kern-.1667em\lower.7ex\hbox{E}\kern-.125emX}}
\begin{document}

\title{Mining SoC Message Flows with  Attention Model}

% \title{A Comparative Study of Specification Mining Methods for SoC Communication Traces\\

\author{ 
% \IEEEauthorblockN{Anonymous}
% \IEEEauthorblockA{Anonymous University\\ Anonymous emails}

% \author{Md Rubel Ahmed, Bardia Nadimi, ~Hao Zheng}
% \affiliation{%
%   \institution{University of South Florida}
%   %\streetaddress{}
%   \city{Tampa}
%   \state{FL}}
% \email{{mdrubelahmed, bnadimi, haozheng}@usf.edu}

\IEEEauthorblockN{ Md Rubel Ahmed, Bardia Nadimi, Hao Zheng}
\IEEEauthorblockA{
U of South Florida, Tampa, FL\\
% Tampa, FL\\
\{mdrubelahmed, bnadimi, haozheng\}@usf.edu}
}

\maketitle
\pagestyle{plain}

\begin{abstract}

High-quality system-level message flow specifications are necessary for comprehensive validation of system-on-chip (SoC) designs. 
However, manual development and maintenance of such specifications are a daunting task.
% Automatic and loss-less reproduction of critical system-level flows such as read, write, update, invalidate, etc. from the SoC execution traces can assist in the efficient analysis, testing, and validation of designs. 
We propose a disruptive method that utilizes deep sequence modeling with the attention mechanism to infer accurate flow specifications from SoC communication traces. 
The proposed method can overcome the inherent complexity of SoC traces induced by the concurrent executions of SoC designs that existing  mining tools often find extremely challenging. We conduct experiments on five highly concurrent traces and find that proposed approach outperforms several existing state-of-the-art trace mining tools.
\end{abstract}


% % \section{Introduction}
% This document is a model and instructions for \LaTeX.
% Please observe the conference page limits. 

% \input{src-introduction}
% \input{src-literature}
% \input{src-background}
% \input{src-tool-clasifications}
% \input{src-selected-tools}
% \input{src-methodology-experiment}
% \input{src-trace-generation}
% \input{src-summary-observation}
\section{Introduction}
\label{motivation}

% Specification mining for system level operations is a rich avenue of research. 
Model inference from execution traces is a popular system analysis, verification, and testing method. An Intellectual Property (IP) block-level communication model or message flow specification is a valuable resource for understanding IP interactions or SoC execution. However inter-IP communication models are not readily available due to complex life-cycle~\cite{patra_validation_wall, valid_soc} an SoC design. Existing pattern or flow mining tools perform poorly on flow specification mining for SoC traces due to the highly concurrent nature of SoC executions. Trace mining for producing high-quality flow specifications becomes extremely difficult when the SoC design complexity increases with adding multiple cores, multi-level shared and private caches, and high-performance interconnects. 
% During the execution of an SoC design, instances of different flows it implements are executed concurrently. 
Traces extracted from SoC communication fabrics are sequences of messages exchanged among IP blocks. A set of well-defined message flows or flow specifications govern the ordering of different messages. An incorrect ordering of the messages could fail an SoC operation. Usually, a flow consists of a set of messages and a unique sequencing rule over the messages. In addition, different flows can have a shared set of messages. Multiple instances of different flows are typically executed concurrently while an SoC is in operation. We primarily focus on mining message flows for complex SoCs. 
% The execution traces becomes overly complicated as multiple tasks can run in parallel in the SoC yielding a very complex traces where finding event or message correlation is nearly impossible. 

The term \emph{message} denotes the unit of exchange between two IPs in the SoC. Though complex in nature, the message flows could be represented as finite state machines (FSM)s. The SoC architectural simulator gem5~\cite{gem5} includes such FSMs to describe its CHI ruby protocol~\cite{muck_2022}. 
\begin{figure}[tb]
\begin{center}
\includegraphics[width=.48\textwidth,angle=0]{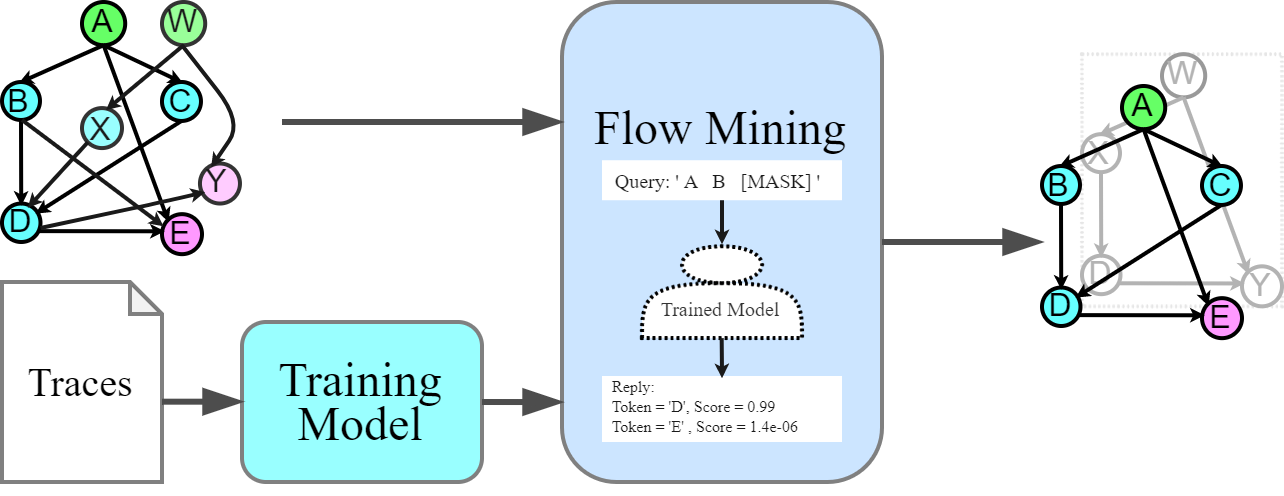}
\vspace{-1pt}
\caption{The overview of Flow Mining from SoC execution traces. Messages from traces are used to build the causality graph. A language understanding model is trained on the traces. Causality graph is then refined with the help of trained model to form the message flow specification.}
\label{fig:workflow}
\vspace{-4mm}
\end{center}
\end{figure}
Specification mining problem is prevalent both in hardware and software domains. Therefore many works have addressed this specification mining problem both from hardware and software perspective~\cite{ahmed2020mining,cao_mining_2020}. We study several tools and techniques for trace mining and find that those tools often fail to find high-quality models. Available approaches fail to extract complex sequencing relations among the messages; therefore, inferred models are often too straightforward or not very interesting to the users. High time and space complexity are also associated with each state-of-art method. For example, a large trace  (>1GB or >7M events) with a high amount of interleaving, \textit{Trace2Model (T2M)}~\cite{natasa2020} can not produce any Deterministic Finite Automata (DFA) within 24hrs. \textit{Synoptic}~\cite{Beschastnikh:2011:LEI:2025113.2025151} reaches the heap memory limit (>24GB) before inferring any system model, and \textit{Texada}~\cite{texada_main}, though very fast, produces a massive amount of linear temporal logic (LTL) rules for the same trace. Comprehending many LTL rules is another challenge; thus, the utility of \textit{Texada} is questionable. Therefore, there is an urgent need for a method to find an execution/communication model from large, interleaved traces within a reasonable time.

After careful study of the SoC traces and recent advancements in sequence modeling tasks, we find that machine learning models for language or sequence understanding can overcome trace length constraints.  Usually, large sequential data or traces can improve the learning of such models.  However, interleaving of flow instances is still a big challenge for finding the correct event or message causation. As the traces are expressed in orderly arrangements of events or messages that follow ground truth flows, the attention mechanism in sequence modeling~\cite{graph2seq} shows a big promise for solving this problem.
This paper presents a framework for addressing the message flow mining from highly concurrent and extensive SoC communication traces.  The overview of our flow mining method is shown in Fig.~\ref{fig:workflow}.  It takes as input a set of execution traces over messages observed in various communication fabrics in an SoC design and produces a set of message flow specifications or flows.
% Input traces from SoC execution go through a preprocessing step that results in building a causality graph where all the participating flows are glued together. 
At first, a lightweight masked language model (LM) trains on the traces to capture the context of the input traces in the form of latent space. The model functions as an oracle or token predictor during a causality graph (CG) refinement process. A causality graph is built from the static information: source and destination tokens of each message. Unfortunately, the straightforward causality graph contains all the flows but many false causal relations among the messages. Therefore, we need some technique to distinguish between correct and false causal relations. The job of the trained language model is to remove such message causality from the graph that lacks minimum contextual relations. The minimum contextual relation is a threshold that the user selects. The higher the threshold, the more accurate the causality in the GC. The refined CG can be transformed into a Finite State Machine (FSM) representation, which could further be used by the tester or validation experts. 

The key feature of this work is that it targets highly concurrent and large-scale communication traces. Furthermore, it utilizes the latest language understanding model to capture the long-term dependency among the messages. This work makes the following {\bf contributions}.
\begin{itemize}
    \item To the best of our knowledge, the proposed method utilizes an attention mechanism for automatically mining SoC message flow specifications. It only depends on the static design information found in the traces.
    \item The proposed method is scaleable to very long traces, as shown in the experiments with realistic SoC traces. 
    \item An evaluation method is also proposed to measure the quality of mined flows. 

\end{itemize}

The organization of this paper is as follows. 
Section~\ref{sec:rel-work} reviews some closely related work in trace mining. 
Section~\ref{sec:background} provides necessary background for related concepts, and formulates the problem. 
Section~\ref{sec:method} describes the proposed trace mining method. Section~\ref{sec:results} presents the experimental results, and the last section concludes the paper.

% \begin{enumerate}
%     \item Language based methods can overcome the trace length constraints
%     \item Attention mechanisms are better than existing LSTMs for finding flows
% \end{enumerate}

% \vspace{-12mm}
% Consequently, an SoC execution trace is a sequence of sets of message instances. The presence of sequential ordering of messages in the SoC traces motivates us for the applying attention mechanism for flow mining. Several works~\cite{yuting_LSTM, ahmed2020mining} try to extract flows in the form of sequential patterns. However, each method comes with some limitations that makes finding complex flows partially possible. The highly proliferating field of attention~\cite{NIPS2017_attention} mechanism to understand language modeling is a good fit for this task where branching in the flows arise. Given enough context, bidirectional encoders can learn the relation among the messages in the traces and predict most suitable message or 'word' to complete the flow. We utilize the lighter version of this bidirectional encoder transformers called distilBERT~\cite{sanh2019distilbert} to reduce the computational and spatial complexity of the method in general.
% \smallskip
% \noindent{\bf Related Work.~}

\section{Related Work}
\label{sec:rel-work}

Message flow mining aims to construct execution models
from various design artifacts. 
\textit{Synoptic} \cite{Beschastnikh:2011:LEI:2025113.2025151} presents an FSA
model-based approach that mines invariants from logs of
sequential execution traces. Generated FSA satisfies the mined invariants.
However, this tool performs poorly when there is a high degree of interleaving in the trace, a common scenario in SoC executions. Model synthesis technique {\it AutoModel}~\cite{AutoModel} extracts concise models from the highly interleaved traces. It can work with very long traces but produces too many models that often miss the true causality of the messages.The work \textit{BaySpec} \cite{Mrowca:2019:LTS:3316781.3317847} produces LTL rules from Bayesian networks trained with software execution traces.
%It is not restricted by some user defined template, thus may produce many useful unknown patterns. 
The techniques presented in in~\cite{Li2010DAC,Hertz:2013:tcad,Danese:2015:vlsi-soc,Danese:2015:date,Danese:2017:dac}  mine assertions from
hardware traces in either gate-level models~\cite{Li2010DAC}, or RTL designs~\cite{Hertz:2013:tcad,Danese:2015:vlsi-soc,Danese:2015:date,Danese:2017:dac}. The work in~\cite{Liu:2013} presents an assertion mining method employing episode mining from the transaction level simulation traces.
A recent work~\cite{Ahmed:mine-msg-flow:2021} addresses a similar flow specification mining problem as in this paper. It captures the temporal relations among the messages as invariants or sequential patterns using the association rule mining technique. However, a lot of complex message relations are not captured in the invariants, which limits the capability of this tool.

Work~\cite{Noc_fabrics} proposes a security property validation method focusing on the communication fabrics of an SoC. It builds a CFG for each IP in the SoC and efficiently explores connections between various CFGs for any given security property. This work considers system-level interaction and demonstrates that communication fabrics could be a vital point for vulnerability. Therefore, understanding the fabric-level communication model is essential for validation. 
Recently, \emph{Trace2Model} is introduced in~\cite{natasa2020} that learns non-deterministic finite automata (NFA) models from platform-agnostic execution traces using C bounded model checking technique. Similar work is also done in~\cite{Ulyantsev:2011}.
Another tool called \texada~\cite{texada_main} works with user-specified templates in the form of LTL and produces instances of that formula using some interestingness measures.
%remain unavailable for many cases in test and debug. 
The above approaches do not consider the concurrent nature of SoC design communication traces and rely on traces' temporal dependencies to identify execution models.
\section{Message Flows, Causality graph, Language Models}
\label{sec:background}
% {\bf check if sequential patterns, support, fw/bw confidences, etc are needed. If not, remove them}

% \begin{enumerate}
%     \item Use FlowMiner to find binary patterns for different f/bw confidence values
%     \item Binary patterns are used to create the cas graph
%     \item We know the start and terminal messages from the cas graph
%     \item We train BERT/LSTM on the traces
%     \item Refine cas using the BERT oracle model
%     \item Transform CAS to FSM
%     \item Use evaluation algorithm to find the FSM acceptance rejection rates for the traces
% \end{enumerate}

{\it Message Flows: } Message flows specify protocols of how system components communicate to implement the system-level functions. For example, high-level operations such as copying a chunk of bytes from a USB device to the hard drive and performing some update on the data can be viewed as a set of small primitive functions that must be executed in a specific order to complete that operation. This ordering of primitive tasks can lead to unique directed graphs: causality graphs (CG) for each system-level function. In a causality graph, each node represents a unique message, and each edge represents causal relations among the nodes or messages. For example, in a multi-core system with multilevel caches and IO peripherals, there could be multiple ways of completing an operation based on the cache-coherence protocols and available peripheral devices, thus yielding multiple branches on each graph. A flow is a collection of messages as nodes and edges as their causal relation expressed as a directed graph. Several previous works~\cite{yuting_prot} have formalized the message flow mining problem in the literature.
% Each message under consideration of this work is a triple $({\tt src: dest: cmd})$ where the ${\tt src}$ refers to the originating IP of the message, {\tt dest} is the destination IP of the message.  Field {\tt cmd} is the operation to be performed at {\tt dest}. Generally, each message flow has a unique {\tt start} message, and a unique {\tt end} message.

\begin{figure*}[tb]
\begin{center}
\includegraphics[width=0.80\textwidth]{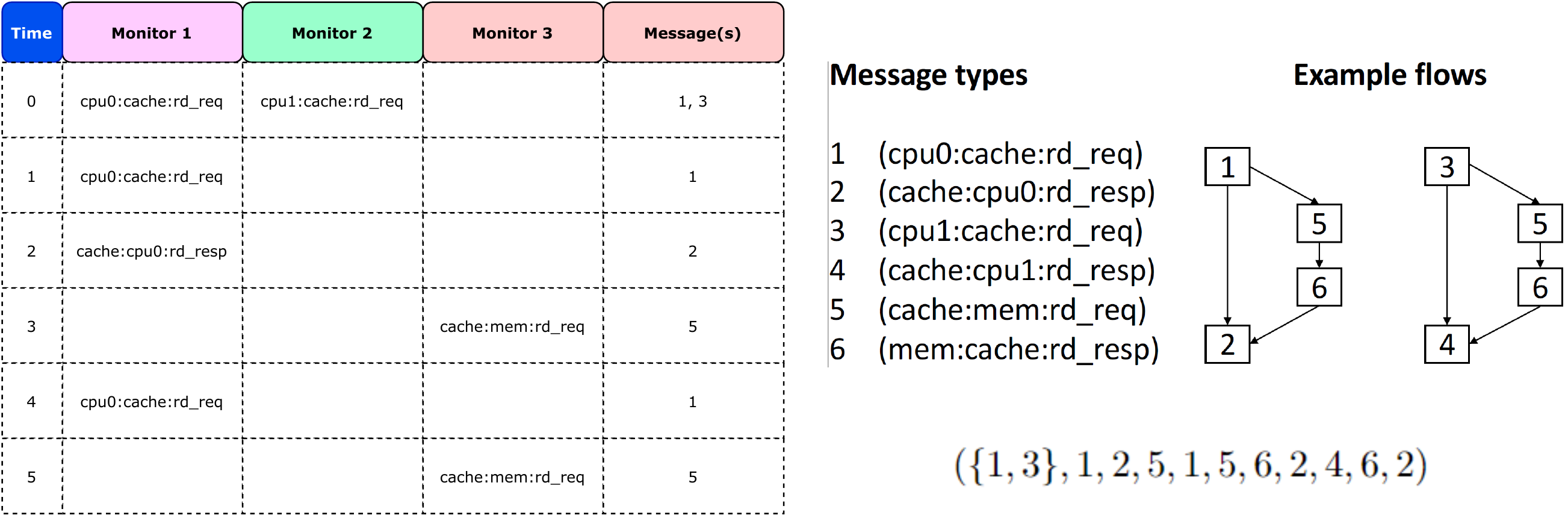}
% \includegraphics[width=.50\textwidth]{figures/message_flow.pdf} & \includegraphics[width=.38\textwidth]{figures/flow-example-simple.pdf}
%\vspace{-2pt}
\caption{A CPU downstream write flow~\cite{AutoModel}.}
\label{fig:flow-ex}
\end{center}
\vspace*{-3mm}
\end{figure*}

A message flow specification is shown in Fig.~\ref{fig:flow-ex}. This simplified example specifies two memory operations via a shared cache for two CPUs. Let us say $M$ is the set of messages that consists of the flow, $C$ is a set of connections among the messages, such that $C \subseteq M\times M$. Each message is a triple $({\tt src: dest: cmd})$ where the ${\tt src}$ denotes the originating component of the message while {\tt dest} denotes the receiving component of the message. Field {\tt cmd} denotes the operation to be performed at {\tt dest}.  For example, message ${\tt (CPU\_0: Cache: rd\_req)}$ is the read request from {\tt CPU\_0} to {\tt Cache}.  Each message flow is associated with a unique {\tt start} message and may terminate with one or multiple different {\tt end} messages. As a message flow may contain multiple branches, Fig.~\ref{fig:flow-ex} has two branches specifying read operations in cases of a cache hit or miss.
The flow specification for an SoC is a set of flows, denoted as $\vec{F} = \{F_i\}$.

During the execution of an SoC design, instances of flows it implements are executed concurrently. When a flow instance is executed along one of its paths, messages on that path are exchanged with runtime information, e.g., specific memory addresses. Multiple instances of different flows executed concurrently are captured in the traces.
%Consequently, an SoC execution trace is a sequence of sets of message instances.
\begin{definition}
%Suppose that an SoC design implements a flow specification $\vec{F}$. 
An SoC execution \textbf{trace} $\rho$ is 
    $\rho = (\varepsilon_0, \varepsilon_1, \ldots, \varepsilon_n)$
where $\varepsilon_i = \{m_{i,0},\ldots, m_{i,k}\}$ is a set of messages observed at time $i$, and $m_{i,j} $ is an message instance of some flow instance active at time $i$, for every $m_{i,j} \in \varepsilon_i$.
\end{definition}

An example trace from executing each path once of the example flows in Fig.~\ref{fig:flow-ex} is 
%where both CPUs initiate a flow at the beginning is  
$\left(1, 3, 5, 1, 5, 6, 2, 3, 6, 2, 4, 4\right)$. Given two messages $m_i$ and $m_j$ and a trace $\rho$, we define $m_i < m_j$, denoting that $m_i$ occurs before $m_j$, if $m_i \in \varepsilon_i$, $m_j \in \varepsilon_j$, and $i<j$. 
%Here $i, j$ can be deemed as two seperate clocks in the SoC execution.

\smallskip
{\it Causality Graphs: }
A causality graph can be constructed from the example flows in Fig.~\ref{fig:flow-ex} if we examine the source and destination information of the message types.  However, such graphs are not easy to construct for large traces. First, assume that the start and end messages are known, then define the causality relation.
% Therefore, we utilize the static information pertaining to a message while building the causality graph using the tool \flowminer. The binary patterns it generates from the traces are used to build the causality graph. Necessary definitions for finding binary patterns goes below:

\begin{figure}[tb]
\begin{center}
\includegraphics[width=.25\textwidth]{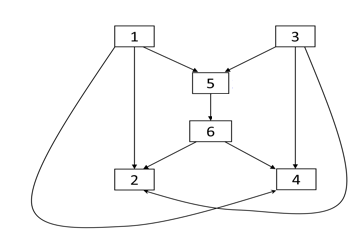}
%\vspace{-2pt}
\caption{Causality graph from example flows in Fig.~\ref{fig:flow-ex}.}
\label{fig:cas_example}
\end{center}
\vspace*{-8mm}
\end{figure}

\begin{definition}
\label{causal}
Messages $m_i$ and $m_j$ satisfy the structural \textit{causality} property, denoted as $\mathit{causal}(m_i, m_j)$, if $m_i{\tt .src} =~m_j{\tt .dest}$
\end{definition}

The causality as defined above is referred to as \emph{structural} to differentiate from the functional causality in the flow specifications. An example trace following the flows in Fig.~\ref{fig:workflow} (top left) could be as follows   
{\tt (A, B, D, W, X, D, Y, E)}, where {\tt A} and {\tt E} are {\tt start} messages while {\tt W} and {\tt Y} are {\tt end} messages, therefore are drawn in different colors. Green nodes are {\tt start} messages and purple nodes are {\tt end} messages. The structural causality property among the messages $m_i$ and $m_j$ is satisfied and denoted as $\mathit{causal}(m_i, m_j)$, if $m_i{\tt .src} = m_j{\tt .dest}.$ The {\tt start} and {\tt end} messages are required to build the causality graph.

\smallskip
{\it Language Models: } Language modeling algorithms have significantly advanced sequence modeling tasks. For example, the self-Attention mechanism~\cite {attention} is a proven technique for capturing long-term dependencies or transducing contextual information in the sequences. This method is heavily used in translation, text classification, and following sequence prediction tasks. Collectively these models are also called transformers. The basic architecture of a transformer contains a stack of encoder and decoder layers. The encoder takes the input texts or symbols and performs some natural language processing such as positional embedding. The decoder employs a masking technique so that the model learns to predict the $i^{th}$ element, relying on the sequence's $(i-1)^{th}$ elements. Such language models are a promising candidate for SoC traces as they also have long sequential dependencies. 
\section{Mining Framework}
\label{sec:method}
%During execution of an SoC design, instances of flows it implements are executed concurrently.
% Typically, multiple instances of different flows that are executed concurrently are captured in the traces. Consequently, a
% An SoC execution trace is a sequence of message instances. 

% \end{definition}

Algorithm~\ref{algo:flowmining} implements the mining framework shown in Fig.~\ref{fig:workflow}.
It consists of three major functions Trace Processing, Training Model,
Flow Mining, which are described below. 

\setlength{\textfloatsep}{6pt}% Remove \textfloatsep
\IncMargin{.5em}
\begin{algorithm}[tb]
\caption{\textbf{Flow Mining}}
\label{algo:flowmining}
\textbf{Input:} a set of traces $\rho$, probability threshold $\theta$ \\ %(also known as score)
\textbf{Output:} {a set of message flows $\mathcal{M}$}\\
\SetAlgoNoLine
Extract unique messages in $\rho$ into $M$\;
Build the causality graph $G$ from $M$\;
Train Model $B$ on $\rho$\;
% $Next$(Start, End, $\theta$) Until $Next$ = $\null$\;
\textbf{foreach} $(Start, End) \in G$ \textbf{do}\\
\hspace{15pt}Query $B$ for next message $m.score() \geq \theta$ towards $End$\;
\hspace{15pt}Return causality subgraph as a $Flow_{(Start, End)}$\;
% Get a solution $sol$ of $P$ using a SMT solver\;
% Derive a FSA $\mathcal{M}$ from $sol$\;
\end{algorithm}
\DecMargin{.5em}

\subsection{Trace Processing}
The proposed framework accepts SoC traces $\rho$ as input. It assumes that each message's source and destination attributes are known, and messages are observed on system communication fabrics and recorded as they occur. Trace processing aims to train the model with highly correlated traces. A language model is trained on the input sequences masking the input by parts. So learning on poorly correlated traces will affect the sequence prediction negatively. While training the model, we employ in-place causality slicing because messages that are not causally related will not form any message flow. Therefore causality slicing increases the message correlation. 

% {\bf unclear!!}~~~
% Trace preparation for training also preserves the causality. The proposed model reads one line at a time, therefore traces are divided into smaller parts or traces $\rho_t$ $\subseteq$ $\rho$.

% \smallskip
% \noindent{\bf Training Model.~~}
\subsection{Training Model}
We utilize a Masked LM {\it DistilBertForMaskedLM} model as the bidirectional transformer. We reuse the default architecture of the transformer and retrain it on our SoC traces to predict masked tokens in a sequence.  We use the light version of {\it BERT}~\cite{devlin2018bert} namely {\it DistilBERT}~\cite{sanh2019distilbert} which contains only 12 encoder layers. This bidirectional model is more compelling than the shallow left-to-right or otherwise models. 
The objective is to train the model to predict the next word in a multi-layered context. To train the model, we randomly mask a certain percentage of the input traces in token form. We apply several optional fine-tuning to fit our traces better.

% \noindent{\bf Flow Mining.~~}
\subsection{Flow Mining}
Message flow mining happens in three steps. The first step is to build the CG from the message {\tt src} and  {\tt dest} information. This CG is supposed to contain all the message flows and additional false branches. We can vary the confidence threshold for our desired complexity of the graph. The next step is to refine the graph.
Once the model is trained, graph refining happens in a guided fashion. There is often incorrect message dependency or causality added to this graph since two messages have a common {\tt dest} and {\tt src} field. For example, in the causality graph Fig.~\ref{fig:cas}~(a), {\tt msg\_19} causes {\tt msg\_27}, though {\tt msg\_27} may not be present in the flow originating with start message {\tt msg\_2} that ends at {\tt msg\_26}.  A search starts from the \emph{start} message and reaches to the \emph{end} message forming a causality sub-graph $G_f$ $\subseteq$ $G$.
The trained model predicts if $m_2$ follows $m_1$ with some probability score returned by method $m_2.score()$.
If the score is greater or equal to the threshold $\theta$, the edge from $m_1$ to $m_2$ is kept in the causality graph. Otherwise, such an edge is discarded. This process is repeated until all flows in the causality graph are discovered for respective $(start, end)$ pairs. 
Finally, the refined CG is transformed into an FSM model. Transformation is straightforward; therefore, the discussion is omitted. Evaluation Algorithm~\ref{algo:evaluationAlgorithm} then evaluates the FSM model against the input traces $\rho$. It reports the number of events rejected by the model. The model with the least number of rejections is reported to the user.

\section{Experiments}
\label{sec:results}
To evaluate the presented framework, a non-trivial SoC
design is simulated in the gem5 architectural simulator, as shown in Fig.~\ref{fig:gem5fs}. gem5 produces a trace that has more than 7M events. The multi-core gem5 simulation contains a high degree of interleaving. Besides a set of synthetic traces taken as a benchmark from work~\cite{AutoModel}. Table~\ref{tab:synthetic_traces} summaries the traces.

% \begin{table}[!t]
% \renewcommand{\arraystretch}{1.3}
% \caption{Trace description.}
% \label{tab:synthetic_traces}
% \vspace{-1pt}
% \centering
% \begin{tabular}{c| c |c c c }
% \hline 
% \bfseries Traces & \bfseries \#Msg & \bfseries\ & \bfseries Length & \bfseries \#States  \\
% \hline 
% \hline
% \multirow{2}{*}{\vtop{\hbox{\strut small}}} & \multirow{2}{*}{22} & \multirow{2}{*}{} & $920$  & 31 \\
%                                             & & & $1840$  & 31 \\
% \hline
% \multirow{2}{*}{\vtop{\hbox{\strut large}}} & \multirow{2}{*}{60} & \multirow{2}{*}{} & $4360$ & 87 \\
%                                             & & & $8720$ & 100 \\
%  \hline
%  \hline
%  GEM5 & 103 &  & $785480$ & 114\\
%  \hline
 
% % \multirow{2}{*}{\vtop{\hbox{\strut multi-message, }\hbox{\strut interleaved}}} & $FlowMiner$ & 114 & 43\% & 18.75\%\\
% % & & & & \\
% \hline
% \end{tabular}
% \end{table} 

% \usepackage{multirow}
\begin{table}[!t]
\renewcommand{\arraystretch}{1.3}
\caption{Trace description.}
\label{tab:synthetic_traces}
\vspace{-1pt}
\centering
\begin{tabular}{c|c|c}
\hline
\textbf{Traces}        & \textbf{\#Msg}      & \textbf{Length} \\ \hline \hline
\multirow{2}{*}{small} & \multirow{2}{*}{22} & 920             \\
                       &                     & 1840            \\ \hline
\multirow{2}{*}{large} & \multirow{2}{*}{60} & 4360            \\
                       &                     & 8720            \\ \hline
gem5                   & 103                 & 7274984          \\ \hline
\end{tabular}
\end{table}
\subsection{General Trace to FSA Evaluation Strategies }
Many works have proposed different methods to evaluate the specification mining methods. For instance, in \cite{jeppu_extended} the proposed method is evaluated by reverse engineering a set of FSAs from their C implementation and the Simulink Stateflow example models have been used as the evaluation dataset.
Loo and Khoo's method \cite{4023976} has also been used for evaluation in \cite{deep_spec_2017}. The method accepts a ground truth and an inferred finite-state automata as input and computes the precision of the inferred FSA based on accepted sentences by the ground truth and the overall number of sentences generated by the FSA.
The evaluation method used in \cite{10.1145/1835804.1835883} is also similar to the previous method. They will calculate the precision and recall values based on the percentage of events of traces that are generated and can be interpreted by the workflows.
In this paper, we chose to evaluate the proposed specification mining method by generating an FSA based on the causality graph. For comparison, we have chosen to compare the proposed method with other state-of-the-art works using synthetic traces and also using the gem5 traces for more realistic results.

\subsection{Results on Different Traces}
For evaluating the proposed flow specification mining method, at first, a finite state automata is created based on the generated causality graph. The corresponding trace will be evaluated using the created FSA. The algorithm for evaluation can be found in algorithm \ref{algo:evaluationAlgorithm}. Also, the achieved results are illustrated in Table.\ref{tab:result_comparision}.

In Algorithm~\ref{algo:evaluationAlgorithm} after generating the FSA based on the causality graph, a search for instances of the generated FSA among trace events will begin. Every single event in a trace file will be checked and if it could be fitted in any stage of an active FSAs, that trace would be marked as accepted. However, if any event is not accepted by any active FSA nor can start a new instance of FSA, it will be marked as unaccepted. Lastly, after exploring all events in a trace file is done, the final acceptance rate will be reported as the output of the algorithm.

\setlength{\textfloatsep}{6pt}
\IncMargin{.5em}
\begin{algorithm}[tb]
\caption{\textbf{Evaluation Algorithm}}
\label{algo:evaluationAlgorithm}
\textbf{Input:} causality graph, traces \\ 
\textbf{Output:} {acceptance rate}\\
\SetAlgoNoLine
Create an FSA based on the causality graph;\\
\textbf{foreach} event in traces \textbf{do}\\
\hspace{15pt}\textbf{if} it was a start event:\\
\hspace{15pt}\hspace{15pt}Create an FSA instance for the start event;\\
\hspace{15pt}\textbf{else if} the event can be fitted in an instance of the FSA:\\
\hspace{15pt}\hspace{15pt}Mark the event as accepted;\\
\hspace{15pt}\textbf{else} :\\
\hspace{15pt}\hspace{15pt}Mark the event as not accepted;\\
\textbf{foreach} instances of the FSA \textbf{do}\\
\hspace{15pt}\textbf{if} the instance is not in the finish state:\\
\hspace{15pt}\hspace{15pt}Mark all of the events in that instance as not accepted;\\
Report the final acceptance rate;\\

\end{algorithm}
\DecMargin{.5em}

\subsubsection{Synthetic Traces}
In order to create a standardized technique for evaluating the accuracy of transformed FSMs, we generated traces synthetically for some of our tests. To synthetically generate traces, we specify a set of patterns permitted to occur concurrently on multiple simulated cores. This concurrency enables realistic interleaving of these patterns, increasing the corresponding traces' difficulty. This difficulty is a function of the numbers and sizes of the patterns, as well as the number of concurrent cores.

Synthetic traces also enable us to test the finite state machine's accuracy by generating traces known to be false. This is essential as it allows us to distinguish an overly accepting finite-state machine from one that accurately models the behavior of valid traces. As the accurate distribution of natural traces is not known, synthetic traces whose behavior is well understood provide essential insight into the effectiveness of {\it BERT} and competing algorithms, as highlighted in Table~\ref{tab:result_comparision}. The runtime contains the training time and dominates the overall time complexity.


\begin{table*}[!t]
\renewcommand{\arraystretch}{1.3}
\caption{Mining Summary: RT = Run Time (in second), DNF = Did Not Finish, Ratio = events accepted by the FSA with respect to rejected events }
\label{tab:result_comparision}
\centering
\begin{tabular}{|l||lll|lll|lll|lll|lll|}
\hline
\textbf{Traces}            & \multicolumn{3}{c|}{small-10}                                                                                                                          & \multicolumn{3}{c|}{small-20}                                                    & \multicolumn{3}{c|}{large-10}                                                    & \multicolumn{3}{c|}{large-20}                                                    & \multicolumn{3}{c|}{gem5}                                                        \\ \hline \hline
\textit{\textbf{Tools}}    & \multicolumn{1}{c|}{Ratio}                                               & \multicolumn{1}{c|}{size}                       & \multicolumn{1}{c|}{RT}   & \multicolumn{1}{c|}{Ratio} & \multicolumn{1}{c|}{size} & \multicolumn{1}{c|}{RT} & \multicolumn{1}{c|}{Ratio} & \multicolumn{1}{c|}{size} & \multicolumn{1}{c|}{RT} & \multicolumn{1}{c|}{Ratio} & \multicolumn{1}{c|}{size} & \multicolumn{1}{c|}{RT} & \multicolumn{1}{c|}{Ratio} & \multicolumn{1}{c|}{size} & \multicolumn{1}{c|}{RT} \\ \hline
\textit{\textbf{Texada}}   & \multicolumn{1}{l|}{\color[HTML]{0000FF} 0.41}                                                & \multicolumn{1}{l|}{231}                        & {2} & \multicolumn{1}{l|}{\color[HTML]{0000FF} 0.39}  & \multicolumn{1}{l|}{231}  & 5                      & \multicolumn{1}{l|}{\color[HTML]{0000FF} 0.08}  & \multicolumn{1}{l|}{231}  & 7                      & \multicolumn{1}{l|}{\color[HTML]{0000FF} 0.13}  & \multicolumn{1}{l|}{231}  & 11                     & \multicolumn{1}{l|}{\color[HTML]{0000FF} 0.35}  & \multicolumn{1}{l|}{1073}  & 5                     \\ \hline
\textit{\textbf{Synoptic}} & \multicolumn{1}{l|}{\color[HTML]{0000FF} 0.92}                                                & \multicolumn{1}{l|}{122}                        & 86                        & \multicolumn{1}{l|}{\color[HTML]{0000FF} 0.92}  & \multicolumn{1}{l|}{122}  & 129                     & \multicolumn{1}{l|}{\color[HTML]{0000FF} 0}  & \multicolumn{1}{l|}{234}  & 127                     & \multicolumn{1}{l|}{\color[HTML]{0000FF} 0.41}  & \multicolumn{1}{l|}{421}  & 865                     & \multicolumn{1}{l|}{DNF}    & \multicolumn{1}{l|}{DNF}   & DNF                      \\ \hline
\textit{\textbf{T2M}}      & \multicolumn{1}{l|}{0.60}                                                & \multicolumn{1}{l|}{90}                         & 4500                      & \multicolumn{1}{l|}{0.63}  & \multicolumn{1}{l|}{45}   & 7235                    & \multicolumn{1}{l|}{DNF}    & \multicolumn{1}{l|}{DNF}   & DNF                      & \multicolumn{1}{l|}{DNF}    & \multicolumn{1}{l|}{DNF}   & DNF                      & \multicolumn{1}{l|}{DNF}    & \multicolumn{1}{l|}{DNF}   & DNF                      \\ \hline
\textit{\textbf{AutoModel}}    & \multicolumn{1}{l|}{\color[HTML]{0000FF} 0.77}                                                & \multicolumn{1}{l|}{52}                         & 23                        & \multicolumn{1}{l|}{\color[HTML]{0000FF} 0.69}  & \multicolumn{1}{l|}{65}   & 68                      & \multicolumn{1}{l|}{\color[HTML]{0000FF} 0.52}  & \multicolumn{1}{l|}{142}  & 626                     & \multicolumn{1}{l|}{\color[HTML]{0000FF} 0.55}  & \multicolumn{1}{l|}{65}   & 1518                    & \multicolumn{1}{l|}{\color[HTML]{0000FF} 0.95}  & \multicolumn{1}{l|}{45}   & 2240                    \\ \hline
% \textit{\textbf{LSTM}}     & \multicolumn{1}{l|}{0.67}                                                & \multicolumn{1}{l|}{118}                        & 2300                      & \multicolumn{1}{l|}{0.63}  & \multicolumn{1}{l|}{136}  & 2300                    & \multicolumn{1}{l|}{0.63}  & \multicolumn{1}{l|}{241}  & 2300                    & \multicolumn{1}{l|}{0.56}  & \multicolumn{1}{l|}{136}  & 2300                    & \multicolumn{1}{l|}{0.48}  & \multicolumn{1}{l|}{327}  & 2300                    \\ \hline
\textit{\textbf{BERT}}     & \multicolumn{1}{l|}{\cellcolor[HTML]{FFFFFF}{\color[HTML]{000000} {\bf 0.95}}} & \multicolumn{1}{l|}{{\color[HTML]{000000} 109}} & 1600                      & \multicolumn{1}{l|}{0.92}  & \multicolumn{1}{l|}{103}  & 1600                    & \multicolumn{1}{l|}{0.96}  & \multicolumn{1}{l|}{103}  & 1600                    & \multicolumn{1}{l|}{0.90}  & \multicolumn{1}{l|}{103}  & 1600                    & \multicolumn{1}{l|}{0.95}  & \multicolumn{1}{l|}{262}  & 4500                    \\ \hline
\end{tabular}
\end{table*}

%\subsubsection{Results on Synthetic Traces}

% \begin{table}[h!]
% \centering
% \begin{tabular}{||c c c c c||} 
%  \hline
%  Algorithm & TPR & TNR & Time & Size \\ [0.5ex] 
%  \hline\hline
%  AutoModel  & 0.41 & 0.95 & 11 days* & 752 \\ 
%  Trace2Model  & 0.96 & 0.11 &  7 hours & 3412\\
%  Synoptic  & DNF & DNF & DNF & DNF\\
%  LSTM  & 0.98 & 0.76 & 29 hours & N/A\\
 
%  \hline
% \end{tabular}
% \vspace{0.2cm}
% \caption{Comparison of Trace2Model with Competing Approaches - Difficulty \textasciitilde 8000}
% \label{table:1}
% \end{table}

\subsubsection{Emulated Gem5 Traces}

\begin{figure}
    \centering
    \includegraphics[width=3.40in]{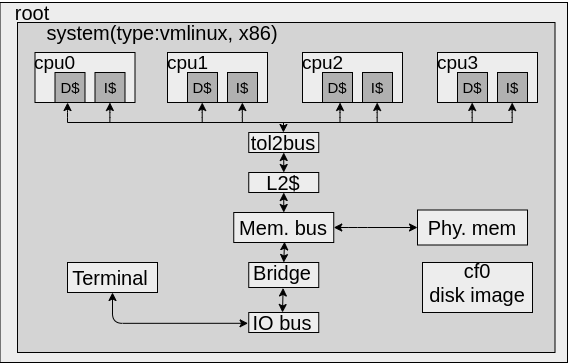}
    \caption{gem5 simulation model}
    \label{fig:gem5fs}
    % \vspace{-3mm}
\end{figure}

To test on a truer-to-life setting, we also generated traces via the use of the \textbf{gem5} emulator. We tracked the flow of messages between roughly a dozen IPs, Fig.~\ref{fig:gem5fs} and logged the flags that were used during those messages to produce a trace of the emulator running various programs, such as fast matrix multiplication and a thread scheduling algorithm. {\it Synoptic} fails to find any model from gem5 traces.

\subsection{Discussion}
\begin{figure}
    \centering
    \includegraphics[width=.32\textwidth]{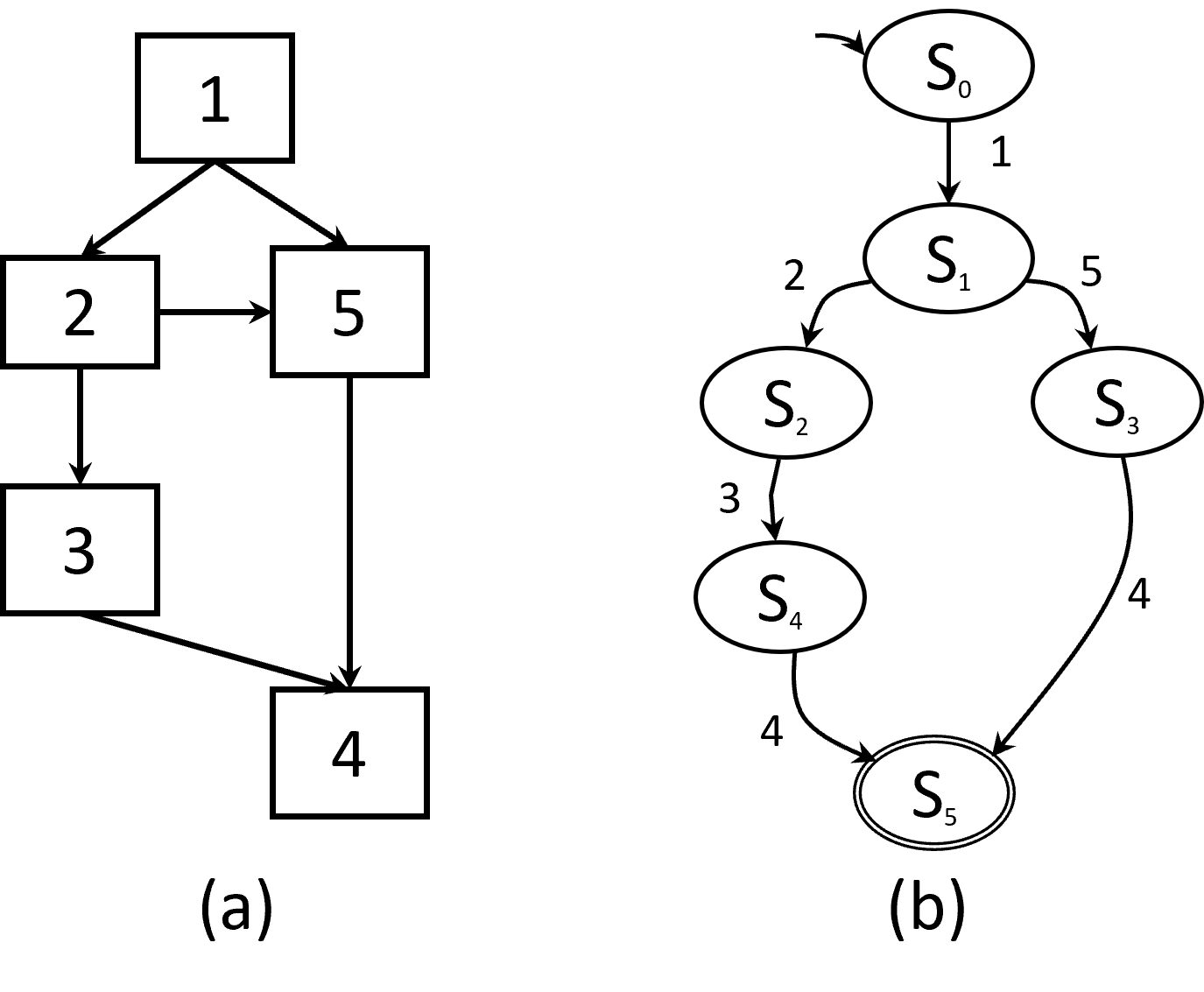}
    \caption{A non-deterministic causality graph (a) and corresponding FSA (b).}
    \label{fig:challengeExample}
    % \vspace{-3mm}
\end{figure}

Using FSAs to validate the traces comes with a challenge. The main one is caused by the non-determinism in the choices of FSA instances. This means that when a new event is being checked with the FSA, there may be more than one possible instance for it to be added to. This situation can be seen in the causality graph in Fig.\ref{fig:challengeExample}. Two of the possible outcomes of following the trace (1, 1, 5, 1, 2, 5, 4, 4, 3, 4) on the causality graph, are \{(1, 5, 4), (1, 2, 3, 4), (1, 5, 4)\} and \{(1, 5, 4), (1, 2, 5, 4), (1)\} which are respectively acceptable and unacceptable (incomplete). Thus, choosing an FSA instance to assign the new event can determine whether the trace will be accepted, and choosing the best FSA instance is the main challenge.

% \begin{figure}
%     \centering
%     \includegraphics[width=3.5in]{figures/conf_size.pdf}
%     \caption{Confidence and size of the solutions}
%     \label{fig:conf_size}
%     % \vspace{-3mm}
% \end{figure}

\begin{figure*}[tb]
\centering

\includegraphics[width=.65\textwidth,angle=0]{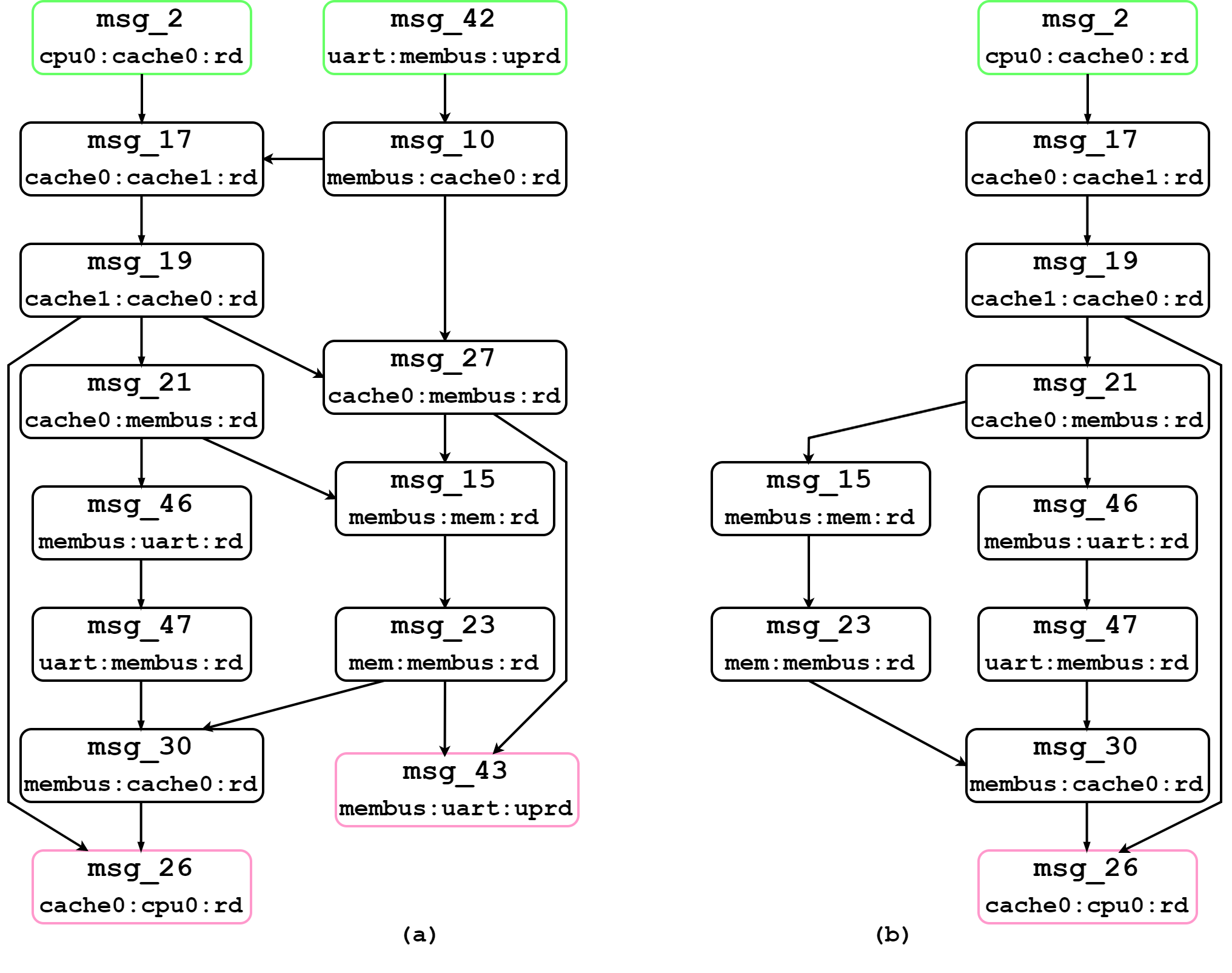}
% \includegraphics[height=3in, width=2.00in]{figures/cas_main.png} 
% \includegraphics[height=.9in]{figures/cas_uart.png}  
% \includegraphics[height=2.2in]{figures/cpu0_read_flow.png}
% \includegraphics[height=2.5in]{figures/uart_upstream_read_flow.png}
% (a)  \hspace{5pt}(b)  (c) (d) (e)
\vspace{-3mm}
\caption{ (a) Causality graph for $\tt CPU0\_Read$ and $\tt UART\_Upstream\_Read$  flows. {\tt CPU0\_Read} flow starts with $\tt msg\_2$ and ends with $\tt msg\_26$ that proposed model refines by removing illegal $\tt msg\_27$. The refinement produces the ground truth $\tt CPU0\_Read$ flow (b) exercised in the trace.}
\label{fig:cas}

\end{figure*}

{\bf Case Study:}
We select two flows $\tt CPU0\_Read$, and $\tt UART\_Upstream\_Read$ adopted from an industry SoC design, which are used to generate a set of traces synthetically. Each flow has three branches, 14 unique messages, and four shared messages, as shown in Fig.~\ref{fig:cas} (a). A training set is prepared to contain 600 concurrent and interleaved runs of these flows. We trained the model for ten epochs, which took around 6.00 hrs. 
% \medskip
% We set the masking a token probability to 30\%.

Fig.~\ref{fig:cas} (a) shows a causality graph built from the messages of the two flows. We want to identify each flow separately with the help of the trained model. A search starts with the $(start, end)$ message pair ({\tt msg\_2, msg\_26}). The branches that do not reach {\tt msg\_26} are discarded in the first place. This gives us a causality graph with many branches to the $end$ message. The trained model helps us remove additional branching edges based on the traces' context learned for each message. For example, $\tt msg\_27$ does not come as a predicted next message for $\tt msg\_19$ for the given score threshold of 75\%. Therefore, we discard message $\tt msg\_27$ from this flow. This process continues till we examine every branch leading to the end message $\tt msg\_26$. Finally, we get the refined causality graph  Figure~\ref{fig:cas} (b), which happens to be the ground truth flow for $\tt CPU0\_Read$. Similarly, we construct $\tt UART\_Upstream\_Read$ flow.

\begin{table}[!t]
\renewcommand{\arraystretch}{1.3}
\caption{Mining results for the three methods on evaluation metrics described in \flowminer~\cite{ahmed2020mining}. $\theta$ = 0.75 for flow mining.}
% \vspace{-3mm}
\label{tab:yuting_lstm}
\centering
\begin{tabular}{c c c c c}
\hline
 &  \multicolumn{1}{c}{ \bfseries \#Patterns Mined } & \multicolumn{1}{c}{ \bfseries Precision} & \multicolumn{1}{c}{ \bfseries Recall} \\
\hline \hline
\flowminer & 53 & 100\% & 25\%\\
$LSTM$ & 22 & 100\% & 3.12\%\\
$BERT$ & $All$ & 94\% & 100\%\\
\hline
\end{tabular}
\end{table}

We apply an LSTM-based pattern mining~\cite{cao_mining_2020} and a rule-based data mining approach~\flowminer on the same synthetic trace set. Work~\cite{cao_mining_2020} extracts flow specification in the form of sequential patterns.
We extract patterns for pattern probability as in ~\cite{cao_mining_2020}, but could not produce a complete specification for any of the tests flows of our experiment. Work~\cite{ahmed2020mining} utilizes association rule mining to find invariants from the traces. The mined rules can partially construct $\tt CPU0\_Read$ where $\tt msg\_15$ and $\tt msg\_23$ are missing. Table~\ref {tab:yuting_lstm} summarizes results for these three methods on the same traces. $All$ refers to the fact that our method does not produce patterns, but DAGs, which could be used to generate all patterns found in the ground truth flows. Regarding evaluation metrics: recall and precision described in~\flowminer, {\it BERT} achieves better scores in both matrices.

% \subsection{Limitations of BERTs}

% Here we discuss the limitations we find for training language models and generating fsms from them. The limitations are as follows and that potentially opens a new venue of research.

% \begin{enumerate}
%     \item Start, end pair is often not found
%     \item Building causality graph could be challenging
%     \item Slicing information might be hard to get
%     \item Hard to find a good evaluation method 
% \end{enumerate}

\section{Conclusion}

We find causality graph-guided word prediction model produces the best quality message flows specifications. However, the high training time could be a limiting factor for this method to existing approaches. Message attribute slicing techniques could help reduce the training time in this regard. Besides, the off-the-shelf models use legacy embedding. An SoC trace-specific embedding could be used further to improve the quality of the mined message flows.
% \input{src-analysis}
% \input{src-conclusion.tex}

% \noindent{{\bf Acknowledgment}
% The research presented in this paper was partially supported by gifts from the Intel Corporation, and a grant from Cyber Florida.

\bibliographystyle{unsrt}
\bibliography{bibfiles/references,bibfiles/soc,bibfiles/others,bibfiles/dm,ref}

\end{document}